\documentclass[conference]{IEEEtran}

\IEEEoverridecommandlockouts
\usepackage{subfigure}
\usepackage{caption}
\usepackage{cite}
\usepackage{amsmath,amssymb,amsfonts}
\usepackage{algorithmic}
\usepackage{graphicx}
\usepackage{textcomp}
\usepackage{xcolor}
\def\BibTeX{{\rm B\kern-.05em{\sc i\kern-.025em b}\kern-.08em
    T\kern-.1667em\lower.7ex\hbox{E}\kern-.125emX}}

\makeatletter
\newcommand*\titleheader[1]{\gdef\@titleheader{#1}}
\AtBeginDocument{%
  \let\st@red@title\@title
  \def\@title{%
    \bgroup\normalfont\large\centering\@titleheader\par\egroup
    \vskip1.5em\st@red@title}
}
\makeatother

\usepackage{import}
\usepackage[]{footmisc}
\graphicspath{ {./images/} }
\IEEEoverridecommandlockouts
\title{Online Advertising Revenue Forecasting: An Interpretable Deep Learning Approach\thanks{This work was funded by Fundação para a Ciência e a Tecnologia (UID/ECO/00124/2019,
UIDB/00124/2020 and Social Sciences DataLab, PINFRA/22209/2016), POR Lisboa and
POR Norte (Social Sciences DataLab, PINFRA/22209/2016)}}
\titleheader{2021 IEEE International Conference on Big Data (Big Data)}

\begin{document}

\author{\IEEEauthorblockN{Max Würfel}
\IEEEauthorblockA{\textit{Nova School of Business and Economics} \\
Universidade NOVA de Lisboa \\
Campus de
Carcavelos, Portugal 2775-405 \\
max.wuerfel@novasbe.pt}
\and
\IEEEauthorblockN{Qiwei Han}
\IEEEauthorblockA{\textit{Nova School of Business and Economics} \\
Universidade NOVA de Lisboa \\
Campus de
Carcavelos, Portugal 2775-405 \\
qiwei.han@novasbe.pt}
\and
\IEEEauthorblockN{Maximilian Kaiser}
\IEEEauthorblockA{\textit{peekd} \\ \\
Berlin, Germany \\
maximilian.kaiser@peekd.ai}
}

\maketitle

\begin{abstract}
Online advertising revenues account for an increasing share of publishers' revenue streams, especially for small and medium-sized publishers who depend on the advertisement networks of tech companies such as Google and Facebook. Thus publishers may benefit significantly from accurate online advertising revenue forecasts to better manage their website monetization strategies. However, publishers who only have access to their own revenue data lack a holistic view of the total ad market of publishers, which in turn limits their ability to generate insights into their own future online advertising revenues. To address this business issue, we leverage a proprietary database encompassing Google Adsense revenues from a large collection of publishers in diverse areas. We adopt the Temporal Fusion Transformer (TFT) model, a novel attention-based architecture to predict publishers' advertising revenues. We leverage multiple covariates, including not only the publisher's own characteristics but also other publishers' advertising revenues. Our prediction results outperform several benchmark deep-learning time-series forecast models over multiple time horizons. Moreover, we interpret the results by analyzing variable importance weights to identify significant features and self-attention weights to reveal persistent temporal patterns.
\end{abstract}


\section{Introduction}\label{intro}
News publishers are continuously challenged by declining print subscriptions and the prevalence of disruptive digital spaces. On the one hand, most adults switch to digital devices to consume news through online channels, such as websites, search engines, and social media. According to a 2020 digital news report published by Reuters Institute, most Americans nowadays choose popular social media, such as Facebook and YouTube, over print media to consume news \cite{reuters2020}. On the other hand, publishers increasingly transform into online publishers and explore new revenue streams from digital subscriptions to online advertisements. A recent survey conducted by Digiday finds that more than half of publishers reported that online advertising was a large or very large source of revenues for them \cite{digiday2019}. This is even more profound for small and medium-sized publishers that depend on the advertisement networks of tech companies such as Google and Facebook. Typically, these news publishers have limited technical sophistication and marketing resources and tend to run digital advertising on their websites through collaboration with advertisement management solutions, such as Google AdSense or Facebook Ads Manager, which together dominate the digital advertising market \cite{emarketer2018}. 

Moreover, online advertising revenues for many publishers have been stagnating since 2015 \cite{econsultancy2015}, in part due to the increasing usage of ad-blocking tools \cite{miroglio2018}, as well as the enactment of consumer privacy regulations regarding behavioral advertising \cite{Marotta2019}. More recently, the concerns regarding advertising revenue declines were further exacerbated by the ongoing Covid-19 pandemic. According to the World Press Trends report published by the World Association of News Publishers, nearly a third of news publishers believe that the decline of advertising revenues represents the biggest threat to the future success of the news industry \cite{Tobitt2021}. As such, given that digital advertising remains a primary source of revenue, publishers may benefit significantly from accurate online advertising revenue forecasts to better manage their website monetization strategies. 

However, publishers that only have access to their own revenues may lack a holistic view of the macroeconomic condition and competitive landscape of the online advertising market. At the same time, all this information is encompassed in the data on the performance of other publishers. This lack of data access limits the ability of any individual publisher to generate insights into their future revenues. For example, although news consumption has experienced massive growth since the pandemic, most online publishers found that advertisers cancel or pause campaigns with them, leading advertising revenues to decrease. Indeed, 69\% of publishers reported that they had to re-forecast advertising revenues due to the pandemic shocks, according to a survey by Internet Advertising Bureau\cite{IAB2020}. 

To address this business question, we aim to forecast online advertising revenues for publishers using a dataset obtained from a German data science company encompassing Google Adsense revenues from a large collection of publishers in diverse areas. We do so by applying deep learning methodologies to perform multi-step time-series forecasting on publishers' advertising revenues over multiple time horizons. In light of the importance of generating interpretable forecasts for business decision-making, we adopt the state-of-the-art deep learning time-series forecasting model Temporal Fusion Transformer (TFT) \cite{49697}, a novel attention-based transformer architecture to predict publishers' advertising revenues. As such, we leverage multiple covariates, including the publisher's own characteristics and other publishers' advertising revenues that may reflect the competitive landscape of the overall advertising market. Our prediction results outperform several benchmark deep-learning time-series forecast models, over multiple time horizons \cite{hochreiter1997long,bahdanau2014neural,DBLP:journals/corr/FlunkertSG17,OreshkinCCB20}. Moreover, we interpret the model predictions using both variable importance weights to identify salient features and self-attention weights to reveal persistent temporal patterns. We find that web traffic unsurprisingly is the most important determinant of publishers' online advertising revenues. Notably, web traffics between publishers is positively correlated in the short-term due to breaking news and negatively correlated in the long-term due to competing sources of similar information, causing correlated advertising revenues. 

This paper offers three major contributions. Firstly, we study the performance projections of publishers at the sell-side of the online advertising industry, while existing literature largely focuses on optimizing advertising systems for advertisers at the buy-side \cite{inproceedings,yuan2013,zhang2014,zhu2017} as well as characterizing consumer behavior\cite{goldfarb2011,gill2013,carrascosa2015}, using the cookie data collected from Internet users. Secondly, we perform time series forecasting on the publisher's online advertising revenues across a range of temporal features from both the focal publisher and other publishers, which has shown as promising industrial applications  \cite{DBLP:journals/corr/FlunkertSG17, bandara2020}. Thirdly, we also identify important features that correlate with publishers' online advertising revenues, which may shed light on their planning and management of monetization strategy. 


The remainder of this paper is organized as follows.
First, Section \ref{related_work} reviews related work on the online advertising market and deep learning models for time-series forecasting, respectively. Next, Section \ref{data} introduces the dataset by discussing the prepossessing steps and the findings from exploratory data analysis. Section \ref{models} details the formulation of time-series forecasting models, while Section \ref{result} presents the results. Lastly, Section \ref{discussion_conclusion} discusses the business insights and managerial implications for the online publishers to maintain sustainable success.

\section{Background}\label{related_work}

\subsection{Online advertising market}\label{online_ad_studies}
The expansion of online display advertising is transforming the advertising industry by providing more efficient methods of matching advertisers and consumers through intermediary platforms, such as Google AdSense and Facebook Ads Manager (see, e.g., \cite{choi2020} for a comprehensive review of history and economic implications of the display advertising ecosystem). For news publishers, online advertising through these systems provides a substantial share of revenues by monetizing their website traffic to advertisers directly or indirectly, without having to manage advertiser relationships \cite{Evans2008}. Advertisers can also reach a relevant target audience based on consumer characteristics and behavior. Typically, publishers use content to attract viewers and create advertising inventory on their websites, such that the provision of advertising is provided by platforms using algorithmic matchmaking between advertisers and consumers \cite{Evans2009,chen2014}. 

More specifically, the ads, which may be rendered on a publisher's website, are automatically selected by the intermediary platform based on three criteria: 1) contextual targeting, which refers to the matching between a website's contents and potential ads;
2) placement targeting, which concerns the preferred consideration of advertisers who registered their intention to advertise on specific websites, 3) personalized targeting, which analyzes the audiences' demographic data, including not only geographical locations and devices but also specific user characteristics, like "sports enthusiast" \cite{googleadsense}. To participate in the advertising network, the publisher only needs to reserve space on their website and inject a code snippet in their web pages. When a viewer visits the page, the publisher will send audience information to the platform, e.g., the location and device and browsing history of the user. Next, the selected advertisers can bid to show their ads in an auction. The ad's rank is not only determined by the cost but also the ad quality score, which tries to capture the expected user experience and the likelihood that the customer will click the ad \cite{googleadsense}. Often, advertisers do not submit bids for individual websites but for keywords, where a keyword may be associated with multiple websites. These keywords represent the contents of the website and relate to the characteristics of the target audience \cite{inproceedings}. The winning advertiser finally sends a script to the publisher's website, where the advertising contents are rendered. Overall, such a real-time bidding process dynamically determines what ads are displayed on the publisher's website, the number of users clicking the ads, and eventually, the revenues earned by publishers \cite{yuan2013}.

The most common online advertising pricing model is expressed in terms of cost-per-click (CPC), which is the amount that an advertiser will pay for a click on its ad to the intermediary platform \cite{Evans2008}. The publishers also have to pay for services to the platform. For example, about 70\% of advertising revenues collected by Google AdSense will be distributed to the publisher \cite{googleadsense}. Adapted from \cite{inproceedings}, a publisher would strive to maximize its online advertising revenues using the following equation with three factors:
\begin{equation*}\label{profitobjective}
    \begin{split}
        Revenue=CPC\times{clicks}=CPC\times{impressions}\times{CTR}
    \end{split}
\end{equation*}
where CTR is the click-through rate, and impressions measure the frequency a specific ad is viewed and are highly correlated with the traffic to the publisher's website. Among three factors, CTR mainly depends on advertising system design and quality of the ad content and does not frequently change, while CPC is mainly driven by advertisers' collective actions, which lies outside of the control of publishers. For example, advertisers might decide to suddenly increase their advertising budget and thereby drive up CPC in an attempt to promote certain products \cite{googleadsense}. Hence, web traffic may account for the most short-term variability in advertising revenues.

Although existing literature involves various predictive methods to estimate CTR \cite{mcmahan2013,yan2014,chen2016,guo2017,zhou2018}, CPC \cite{inproceedings,zhu2017} and web traffic \cite{piramuthu2003,wang2005,li2008forecasting,petluri2018}, respectively, research on online advertising revenue prediction for publishers is largely underexplored. We believe that this stream of research remains scant due to the unwillingness of publishers to disclose their advertising revenues, especially to the competitors. Practically, advertising revenue calculators are available for the publishers to receive a rough estimate of expected online advertisement revenues based on the content category, region, and web traffic. Notably, these key inputs would serve as an indication of expected advertising revenues. However, none of these tools leverage historical information of publishers to generate individualized forecasts but rather yield an industry averaged estimation.

\subsection{Deep learning for time-series forecasting}\label{deep_learning_time_studies}
Deep learning, while gaining its original popularity from Computer Vision and Natural Language Processing (NLP) tasks, has been frequently applied to generate time-series forecasts.
Driven by the increasing data availability and computing power, deep learning allows to learn temporal patterns without the need for extensive preprocessing, assumptions about the underlying distribution and exhaustive feature engineering \cite{lim2020time}.
However, statistical models like ARIMA continue to perform distinctively better on univariate problems, especially for short time series.

In our work, the time series contains multivariate features and spans over a thousand daily observations. With this length of time series data, it has been shown that deep learning methodologies will outperform statistical models reliably \cite{cerqueira2019machine}.
A more practical reason in favor of deep learning methodologies is that only a single model needs to be fit, which can make predictions across all cross-sectional units in the dataset.
Generally, most statistical models show diminishing performance in high-dimensional multivariate problems, where deep learning models start to benefit from their ability to fit non-linear, complex functions.

Furthermore, to make multi-step predictions, models like ARIMA can solely be used recursively, making one-step-ahead predictions, while deep learning models allow for more advanced output strategies.
Finally, compared to statistical methodologies, there is no need to select from parametric and non-parametric models. 
For many statistical models, the data needs to be stationary, such that mean and variance are constant over time.
By contrast, deep learning models usually learn to deal with seasonalities and trends independently. Historically, a caveat to deep learning methodologies has been the lack of interpretability.
However, as new deep learning methodologies emerge, efforts have been made to combat this issue \cite{lim2021}.

In business applications like the one at hand, interpretability of the prediction, i.e., to understand both how and why a model makes a certain prediction, becomes increasingly imperative in order to facilitate business decision support. This study aims to apply deep learning methodologies to generate forecasts, with a special interest in the interpretability of the results.

In many industrial applications, it is preferable to generate a sequence of forecasts over a specified time horizon. For example, publishers would benefit from the forecast of advertising revenues in the upcoming week or month for revenue planning and management purposes.
Multiple strategies exist to make multi-step predictions.
First, a distinct model can be developed for each timestep, each generating its own one-step-ahead prediction. 
As the number of models is linearly increasing in the forecasting horizon, this method is infeasible for longer time horizons.
Second, iterative (autoregressive) deep learning architecture have been proposed to generate multi-horizon predictions. 
Thereby, a one-step-ahead model is trained and at runtime, the one-step-ahead predictions are appended to the input vector. 
The caveat of this prediction strategy is that when using multi-variate inputs, a prediction needs to be made for each input series independently to be able to generate one-step-ahead predictions.
Additionally, errors made in early predictions can accumulate over the prediction horizon \cite{lim2021}.
Finally, it is possible to utilize a multi-output strategy, where a single model outputs multiple predictions at the same time.
While only a single model needs to be trained, this strategy usually requires the usage of more complex model architectures to successfully make predictions.
To directly generate output vectors, basic models like Recurrent Neural Networks (RNNs) can be manipulated to output a fixed-length vector instead of a single prediction. Alternatively, encoder-decoder architectures such as sequence-to-sequence models directly take all available inputs to summarize past information with an encoder and combine them with known future inputs with a decoder \cite{fan2019}.

\section{Problem Formulation and Data}\label{data}
The goal for this paper is to forecast daily AdSense revenues for publishers over multiple days, given their historical AdSense revenues together with a set of time-varying and time-invariant features. More formally, the multi-horizon multivariate prediction problem is defined as:
\begin{equation}
    y_{i,t+\tau} = f\big(y_{i,t-k:t}, u_{i,t-k:t+\tau}, x_{i,t-k:t}, s_i\big)\}
\end{equation}

where target variable $y_{i,t}$ denotes AdSense revenues for publisher $i$ at day $t$, and thus $y_{i,t-k:t}=\{y_{i,t-k},\dots,y_{i,t}\}$ includes historical AdSense revenues from the past $k$ days. Moreover, time-varying features include $u_{i,t-k:t+\tau}=\{u_{i,t-k},\dots,u_{i,t+\tau}\}$, i.e., those that are \textit{known} from the past $k$ days until the future $\tau$ days , such as day of the week and $x_{i,t-k:t}=\{u_{i,t-k},\dots,u_{i,t}\}$, i.e., those that are \textit{unknown} over the prediction time horizon, such as number of daily page views, bounces, sessions, as well as ad impressions and clicks. Lastly, $s_i$ denotes time-invariant features such as the country and category of the publisher.

The data for this study was provided by peekd, a German data science company, including daily Google Adsense revenues from a large collection of news publishers in diverse content categories, spanning from January 2018 to December 2020. The company offers publishers the opportunity to receive anonymized benchmark analytics based on its proprietary database. Additionally, time series for ad impressions, ad clicks, page views, and bounces are available with the same granularity.
As daily revenues are measured across three device categories, namely desktop, mobile, and tablet, we only use the daily advertising revenues generated from the desktop as it captures the largest fraction of the total revenue, and ad clicks and impressions from different device categories convert into AdSense revenues at significantly different rates.

Given that all deep learning models deployed in this work take as input a sequence of daily observations, following the sliding window technique, e.g., deployed by \cite{vafaeipour2014application}, the inputs are constructed as two-dimensional matrices with the multivariate features on one axis and the input time steps on the other axis ($k_{max}=89$).
Hereby, $k_{max}$ was chosen after inspecting AdSense revenue autocorrelation plot (Fig. \ref{fig:acf}a).
Similarly, when $\tau_{max}=30$, the target is given by a vector of 30 univariate observations (we also train models for $\tau_{max}=7$ and $\tau_{max}=14$).
To generate the next sample, both the input and output windows are shifted one day forward.
In time series forecasting, train, validation, and test splits need to be generated chronologically to ensure that all training data stems from dates occurring before any date which is part of the validation and test sets. In this study, only observations before April 2020 are used for training, while data until June 2020 is used for validation, and all remaining dates are used for testing.
\newline

\subsection{Preprocessing}\label{preprocessing}
The dataset encompasses data of over anonymized 400 publishers, and we take a subset based on three criteria. Firstly, the dataset has been reduced to incorporate only publishers from fifteen developed countries, including Australia, Belgium, Canada, Denmark, Germany, Italy, Japan,  Netherlands, Norway, Portugal, Spain, Sweden, Switzerland, United Kingdom, and United States, because the potential AdSense revenue and its determinants CTR and CPC might behave similarly within the socio-economically close countries. Admittedly, the origin of a publisher will not entirely match the origin of its viewers, but it still serves as a reasonable proxy. Another reason for taking a regional subset of the data is that culturally close countries tend to experience similar trends and patterns in their website traffic as well as the overall online demographics.

Furthermore, there are a high number of publishers with constant zero AdSense revenues for a long period in the dataset. One possible reason might be the low barriers of entry, as signing up to Google AdSense require little technical effort and is free of charge. Consequently, even publishers with little website traffic are encouraged to sign up for the service. Unlike handling missing observations, imputation does not seem to be a viable option for such cases, because it is difficult to distinguish them from those publishers who discontinued their usage of the AdSense network. Moreover, time series with too many zero-valued observations would lack variability over the time horizon, such that no trends or seasonality can be observed, which makes it harder for any models to learn meaningful patterns. Last, different sized publishers may have different revenue patterns, e.g., many small publishers experience significantly more volatile revenue patterns than more established ones. As such, similar to preprocessing measures by \cite{rebortera2019enhanced}, publishers with more than ten consecutive zero observations are removed, and only publishers with more than \$5 average daily revenues are retained. As such, 42 publishers remain in the final dataset.

Moreover, both \cite{DBLP:journals/corr/FlunkertSG17}, and \cite{gill2013} reveal the distribution of the revenue time series across the different publishers is highly skewed, which would make standardization techniques like normalization less effective. To alleviate this issue, all daily AdSense revenues are log-transformed, followed by the standard scaling applied to each publisher separately. Thus, all revenue time-series are aggregated by publisher, and then each time series is individually normalized to zero mean and unit variance.

\subsection{Explorative data analysis}\label{eda}
\begin{figure*}[!htb]
    \centering
    \subfigure[AdSense Revenue Autocorrelation Plot]{\includegraphics[width=0.48\textwidth, height=3.5cm]{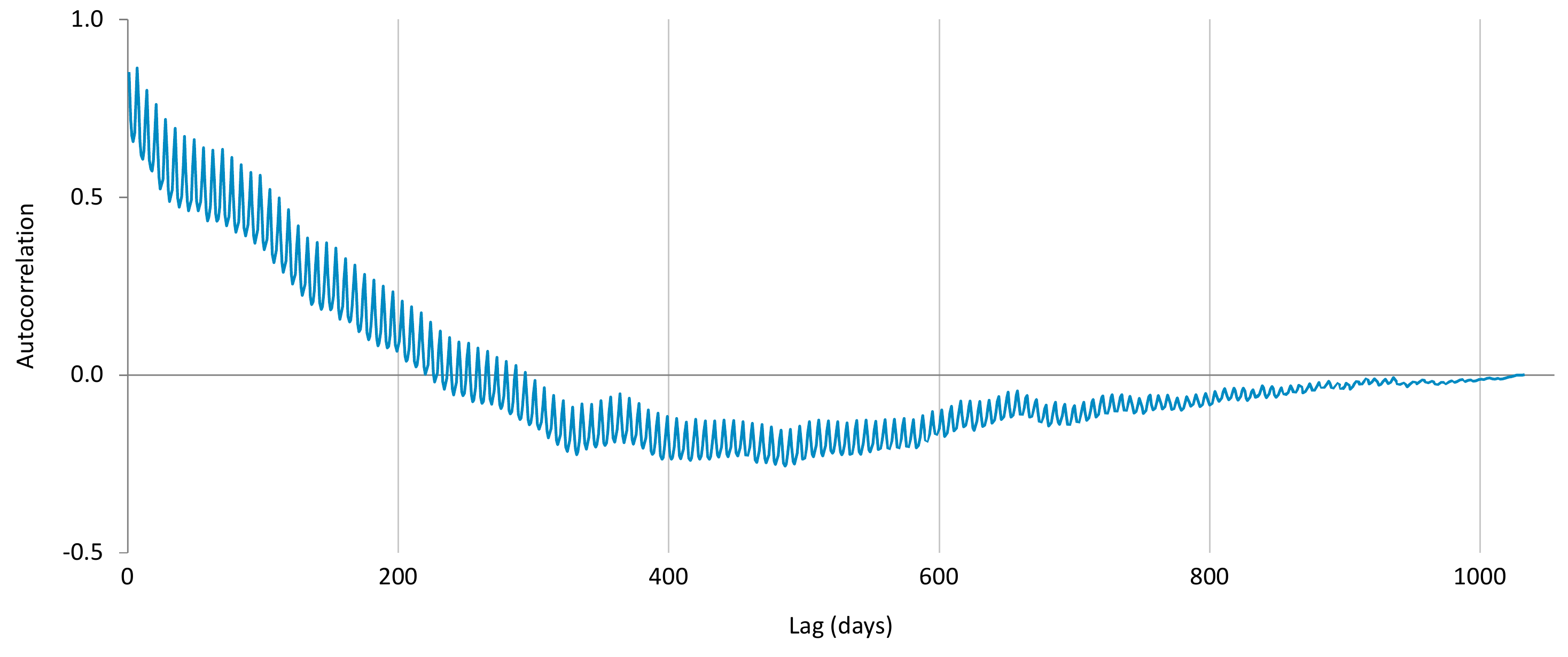}}
    \subfigure[Page Impressions Autocorrelation Plot]{\includegraphics[width=0.48\textwidth, height=4cm]{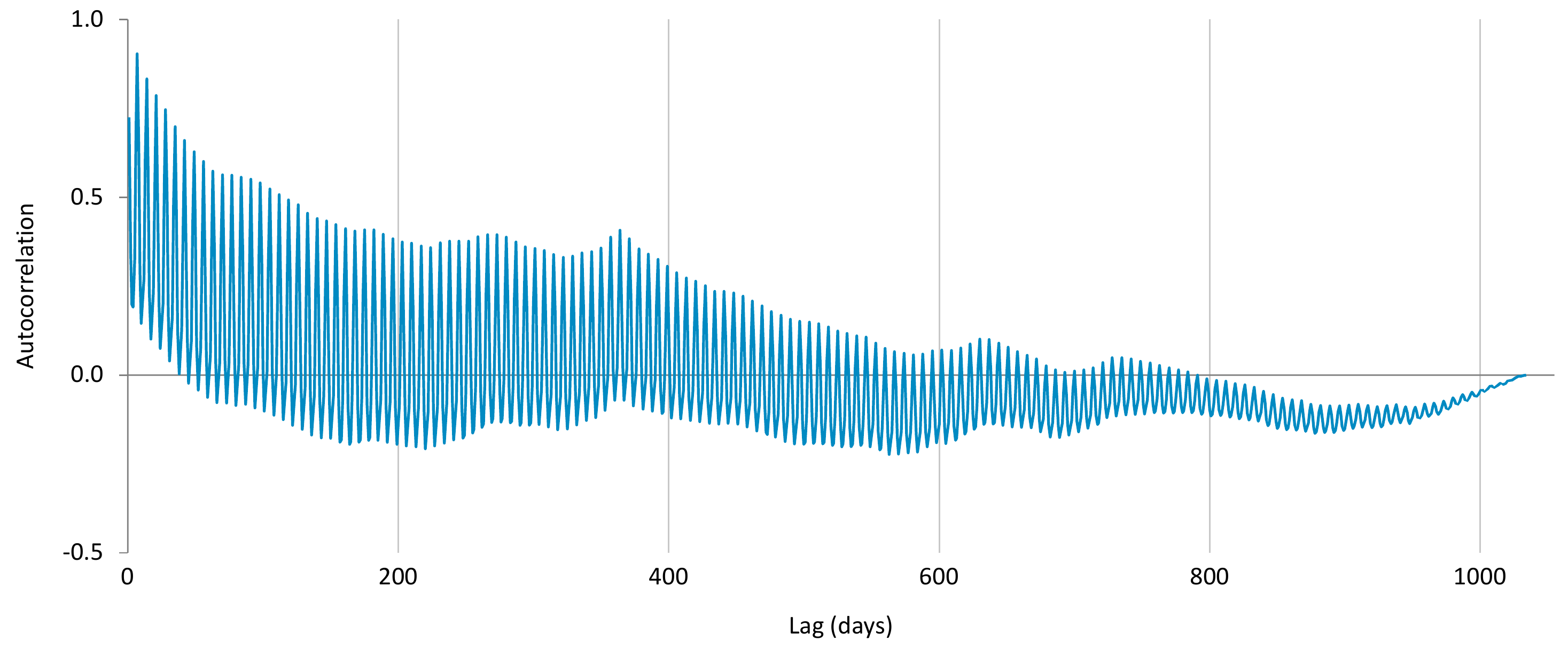}}
    \subfigure[Click-Through-Rate Autocorrelation Plot]{\includegraphics[width=0.48\textwidth, height=4cm]{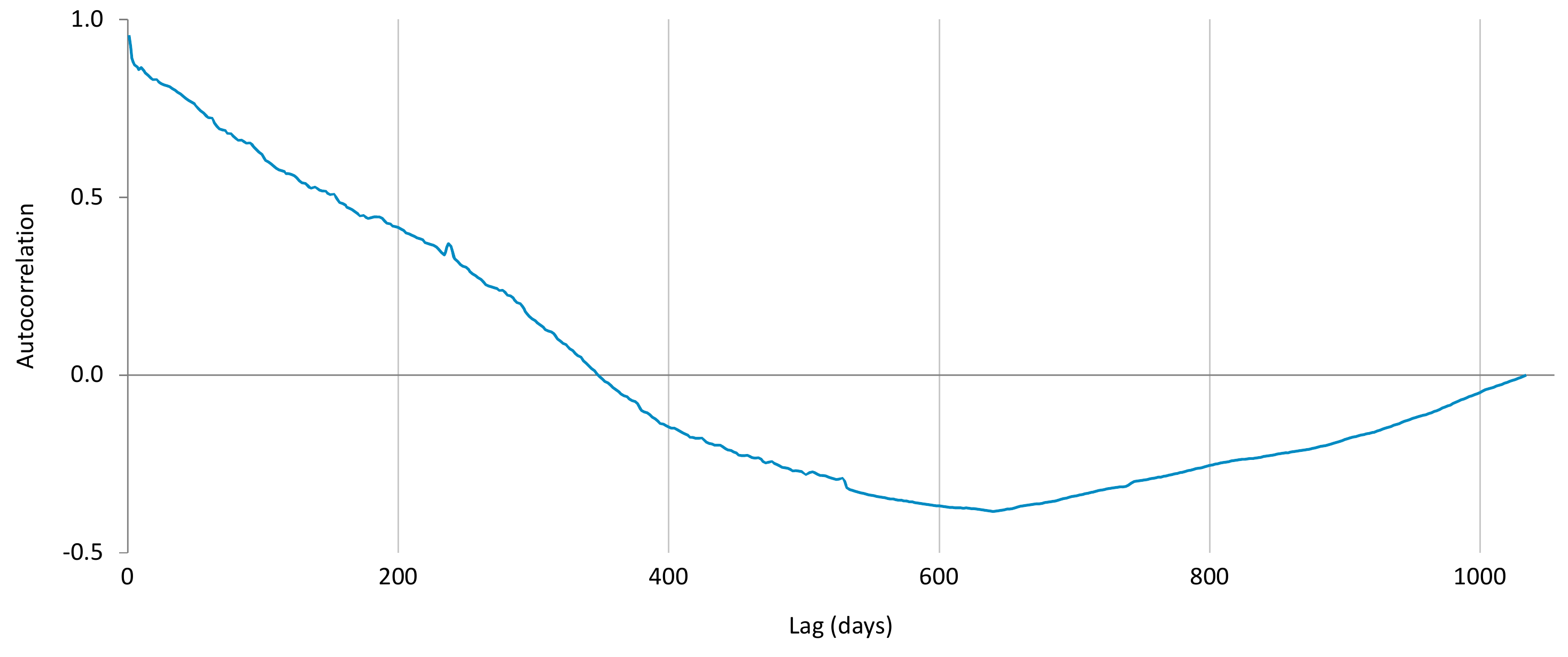}}
    \subfigure[Cost-Per-Click Autocorrelation Plot]{\includegraphics[width=0.48\textwidth, height=3.5cm]{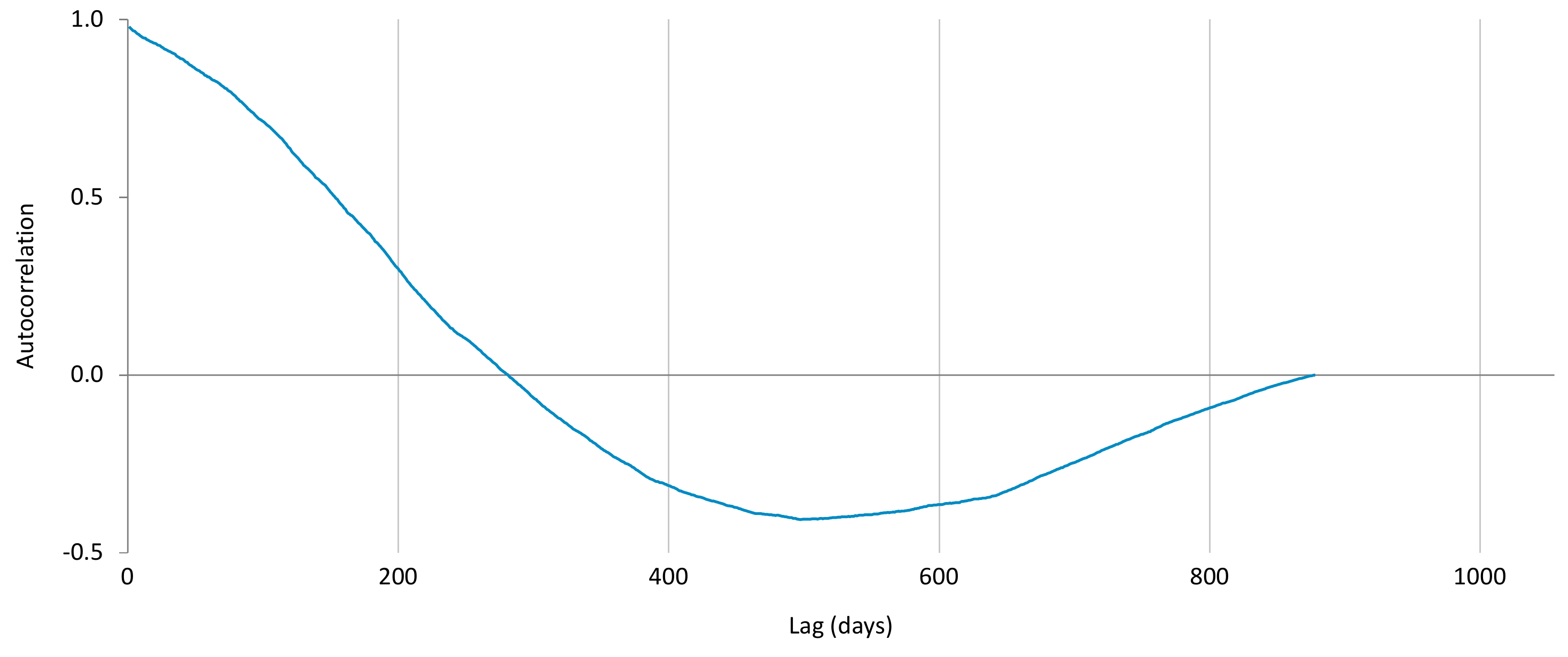}}
    \caption{Autocorrelation plots for key website metrics across all publishers}
 \label{fig:acf}
\end{figure*}

We constructed four autocorrelation plots of AdSense Revenue, impressions, CTR, and CPC in Figure \ref{fig:acf}, respectively, and find preliminary evidence to support the hypotheses discussed in Sec. \ref{online_ad_studies}, that web traffic is likely to have the most significant effect on short-term AdSense revenue patterns. In particular, the autocorrelation plot of AdSense revenues showcase similar autocorrelation patterns (Fig. \ref{fig:acf}a) as that of ad impressions (Fig. \ref{fig:acf}b) that serves as the proxy for web traffic. 
However, autocorrelation plots of CPC (Fig. \ref{fig:acf}c) and CTR (Fig. \ref{fig:acf}d) do not exhibit similar short-term autocorrelation patterns as that of AdSense revenue, but are rather inherently similar between the two. 


\section{Time-series forecasting models}\label{models}
\subsection{Temporal Fusion Transformer (TFT)}\label{tft}
The TFT by \cite{lim2020temporal} consists of multiple processing layers stacked on top of each other.
The first layer is made up of multiple variable selection networks (VSN), where all known, unknown, and static inputs are preprocessed separately by dedicated VSNs.
Importantly, variable selection takes place at each time step separately. 
This allows for increased interpretability through computation of feature importances, while at the same time, features which fail to contribute to model performance will be filtered out.

The next layer performs locality enhancement and is essentially made up of a Seq2Seq LSTM model. 
This is important, as the relevance of individual observations often only becomes visible within their context. 
A typical example is a change point, which can only be identified in context. 
Hereby, the encoder LSTM is fed the known and unknown time-series, while the decoder LSTM is only fed the known time-series, both of which have previously been processed by the corresponding VSNs.
In addition, the encoder LSTMs are hidden and cell states are initialized by the outputs of a VSN, which is fed the static features. 
As in most Seq2Seq architectures, the decoder LSTMs hidden and cell state are initialized by the encoder LSTM's last cell's hidden and cell states. 

The adjacent layer is fully committed to static enrichment.
Specifically, the outputs from the locality enhancement layer are all fed to a gated residual network (GRN). 
The GRN is an essential part of the TFT architecture and is also part of every VSN. 
It is built on the two ideas of gated activations and residual connections, from Sec. \ref{deep_learning_time_studies}.
The intention is to provide the model with the freedom to decide on the amount of required non-linear processing for every input separately. 
When no further complexity is required, no non-linear transformation at all can be applied. Besides a second optional input, the context vector can be supplied to the network.
In the case of the static enrichment layer, the static inputs are fed as the optional input and can therefore influence the processing of the time-series features.
Beforehand, all static inputs have once again been processed by a dedicated VSN.

Next, the temporal self-attention layer allows the model to capture long-term trends by accessing information from the far past. Compared to the original multi-head attention layer from \cite{vaswani2017attention}, TFT made a minor modification to the architecture of the attention heads, which in turn enables the exploration of temporal attention patterns. Basically, while the weight matrices for the queries and keys are learnable and specific to each head, the weight matrix for the value vector is shared across all heads. Thus, each head can still learn distinct temporal patterns while attending to the same inputs, and overall attention can be computed across all heads and subsequently be used for the purpose of interpretation.
Notably, the attention layer will also apply forward masking, such that causal information flow is retained. A position-wise feed-forward layer constitutes the final processing step.
At each forecasting step, a GRN is applied to the outputs of the temporal self-attention layer, with the weights shared across all time steps. 
Further, a gated residual connection allows the outputs from the locality enhancement to skip the entire transformer.

Overall, the TFT model may allow for good predictive performance while ensuring a high degree of interpretability.
First, it applies the transformer architecture, which proved to be successful in time-series forecasting while incorporating a Seq2Seq structure, ensuring locality enrichment.
Second, the distinct handling of the different input variable types might offer an edge over any architecture which repeats static input vectors across time-steps or is unable to handle any of the input variable types.

\subsection{Benchmark models}
\subsubsection{LSTM}
The LSTM acts as a baseline model where each historical time step represents a network cell \cite{hochreiter1997long}.
As each cell corresponds to a historical time step, any cell can be fed the known and unknown time series.
The static features are omitted and not supplied to the network.
The multi-step output vector is computed from the final hidden state, which is fed through a linear layer to match the required output steps.

\subsubsection{Seq2Seq LSTM with Bahdanau (additive) Attention}\label{Seq2Seq}
The most natural choice for a benchmark to the TFT is a Seq2Seq model with a regular attention layer as proposed by \cite{bahdanau2014neural}.
Next to the VSN, the main contribution of the TFT is the transformer structure stacked on top of the locality enrichment layer.
Thus, the TFT should prove to outperform a regular attention layer stacked on top of a Seq2Seq LSTM model.
The Seq2Seq model implemented in this paper consists of two separate LSTMs, one acting as encoder and one as the decoder. 
While the encoder is supplied with all unknown and known time-series, the decoder is fed the unknown time-series, the static features, which are repeated for each prediction time-step, and the context vector.
The context vector is computed at each prediction time step as the product of the encoder hidden states and the corresponding attention weights.
The attention weights are computed using a tanh function, which takes as input the corresponding encoder hidden state as well as the previous decoder cell's hidden state.
Finally, all attention weights are weighted such that they add up to one.
The hidden and cell states of the decoder are both initialized using the hidden and cell states from the last encoder cell.
Different from the TFT, the decoder of the Seq2Seq LSTM works recursively.
To make a prediction, the prediction from the previous time step is concatenated to the decoder inputs.

\subsubsection{DeepAR}
The DeepAR model is an autogressive multi-layer RNN leveraging LSTM cells \cite{DBLP:journals/corr/FlunkertSG17}. Basically, it introduces the conditioning and the prediction range, which can also be thought of as the encoder and decoder in a regular Seq2Seq model like in Sec. \ref{Seq2Seq}.
Any prediction is conditioned on the real values from the (historical) conditioning range as well as any previous predictions for the prediction range.

Although both the encoder and decoder could have different architectures, our implementation of the model treats the architectures the same, and even the weights are shared across encoder and decoder.

Obviously, any unknown time series would only be known for the conditional range and not for the prediction range.
But as explained previously, the architecture and the weights are shared across the conditioning and prediction range. Thus we cannot omit any unknown time-series in the prediction range if we feed it to the model in the conditioning range.
Therefore, the DeepAR model only takes the known time-series as well as the categorical features.

Furthermore, it is important to point out that DeepAR is a likelihood model, such that for real-valued targets, as in the case at hand, the network predicts all parameters (e.g., mean and standard deviation) of the probability distribution for each time step. 
Consequently, it is also possible to generate samples through ancestral sampling at each time step.
To do so, the autoregressive loop runs multiple times, and at every iteration, at each prediction time step a prediction is sampled from the suggested distribution \cite{DBLP:journals/corr/FlunkertSG17}.
Notably, all static features are embedded into vectors and concatenated with the known time series before they are fed to the corresponding LSTM cells.

\subsubsection{N-Beats}
N-Beats is a recent deep neural architecture based on backward and forward residual links, and a very deep stack of fully connected layers \cite{OreshkinCCB20}. N-Beats adopts an expressive yet interpretable architecture and does not require any time-series-specific feature engineering. Its architecture involves two configurations: 1) the generic architecture with a multi-layer fully connected network and 2) the interpretable architecture that incorporates the trend and seasonality decomposition into the model. The forecasts are aggregated using ensemble models in a hierarchical fashion, which enables building a very deep neural network with interpretable outputs. N-Beats has been shown to outperform many statistical models on heterogeneous benchmark time-series datasets. However, N-Beats only supports univariate inputs, which only yield interpretations by decomposing time-series into trends and seasonality. This would disallow us to provide more meaningful business insights given our rich data to characterize publishers' advertisement revenues.

\section{Results}\label{result}
\subsection{Performance comparison}
\begin{table*}[!htb]
\begin{center}

        \begin{tabular}{p{1.3cm}p{1.3cm}p{1.3cm}p{1.3cm}
        p{1.3cm}p{1.3cm}p{1.3cm}p{1.3cm}p{1.3cm}p{1.3cm}}
        \hline
          Model  & \multicolumn{3}{c}{7 days}  & \multicolumn{3}{c}{14 days} &
            \multicolumn{3}{c}{30 days} \\
            \cline{2-4}  \cline{5-7}  \cline{8-10}
             & MAE & MAPE & SMAPE 
            & MAE & MAPE & SMAPE 
            & MAE & MAPE & SMAPE \\
            LSTM  & $15.822$ & $0.759$ & $0.381$ 
            & $17.225$ & $0.648$ & $0.402$
            & $23.992$ & $2.599$ & $0.532$\\
            DeepAR & $11.993$ & $0.474$ & $0.340$ 
            & $14.565$ & $0.704$ & $0.377$
            & $16.583$ & $0.607$ & $0.396$\\
            Seq2Seq  & $11.878$ & $0.358$ & $0.349$ 
            & $14.346$ & $0.451$ & $0.371$
            & $15.821$ & $0.465$ & $0.384$\\
            NBeats & $11.326$ & $0.352$ & $0.324$
            & $\textbf{12.476}$ & $\textbf{0.371}$ & $\textbf{0.338}$
            & $14.841$ & $0.417$ & $0.357$\\
            TFT & $\textbf{11.144}$ & $\textbf{0.352}$ & $\textbf{0.320}$ 
            & $13.150$ & $0.391$ & $0.343$ 
            & $\textbf{14.773}$ & $\textbf{0.416}$ & $\textbf{0.356}$ \\
            \hline
        \end{tabular}

\caption{Model performance comparison between TFT and benchmark models in three horizons (7 days, 14 days and 30 days)}
\end{center}
\label{table_results}
\end{table*}

All models are trained and fine-tuned on the train and validation set and evaluated on the test set using three metrics: 1) Mean Absolute Error (MAE); 2) Mean Absolute Percentage Error (MAPE), and 3) Symmetric Mean Absolute Percentage Error (SMAPE). The mean AdSense revenue of a publisher's content category is supplemented into the multivariate deep learning model as an additional covariate. As \cite{DBLP:journals/corr/FlunkertSG17} suggested, using data from related time series allows complex forecasting models to alleviate the overfitting and manual effort for feature selection. 

Hyperparameter tuning was conducted using the Bayesian optimization framework Optuna, where due to resource constraints, each model was given 20 trials with five epochs each \cite{akiba2019}. Even though the number of epochs might appear small, each epoch works on all samples generated using a one-step forward sliding window. As such, there will be significant
overlap of the daily observations visited across samples. Therefore, the models will converge already after a very low number of epochs. For the TFT, hidden size, dropout, and number of heads were tuned. In the Seq2Seq architecture, hidden sizes in encoder and decoder, respective dropout rates, and the number of layers in the encoder were tuned. Finally, in the LSTM, hidden layer size, number of layers, and dropout were optimized. The batch size was set to 32 and the Adam optimizer was used for all models. The learning rate was only tuned from a subset of 0.001 and 0.0001, all in an effort to reduce hyperparameter space.

Table I shows the model performance comparisons between TFT and several benchmark models in three horizons (7 days, 14 days, and 30 days). The TFT outperforms the other models across all metrics, except to NBeats in 14 days, while providing opportunities to interpret the results.
In particular, the performance of the TFT is distinctively better when compared to the LSTM, the Seq2Seq LSTM, and the DeepAR model, and although it only marginally outperforms the NBeats model, TFT model may generate interpretable predictions which may shed light on the publisher's revenue management.

Moreover, as expected, we observe that the performance of all models deteriorates with increasing prediction time horizons.
Most significantly, we find that adding the mean AdSense revenue of a publisher's category as an additional covariate has helped to improve model performance. 
Consequently, we have collected evidence indicating that it is possible to enhance model performance through access to a cross-sectional database and the specific selection of correlated time series.
\cite{DBLP:journals/corr/FlunkertSG17}, for example, show that models trained on cross-sectional data can outperform models learned from a single unit time series simply based upon the ability of deep learning models to learn generalizable functions.
We add to these findings by showcasing that given a suitable business scenario, where the revenues of different units are necessarily correlated, the performance of multi-variate time-series prediction models, like the TFT, can be increased through careful selection of correlated time-series.

\begin{figure*}[!htb]
    \centering
    \subfigure[TFT  - Top 1 prediction (30 days)]{\includegraphics[width=0.48\textwidth, height=3.5cm]{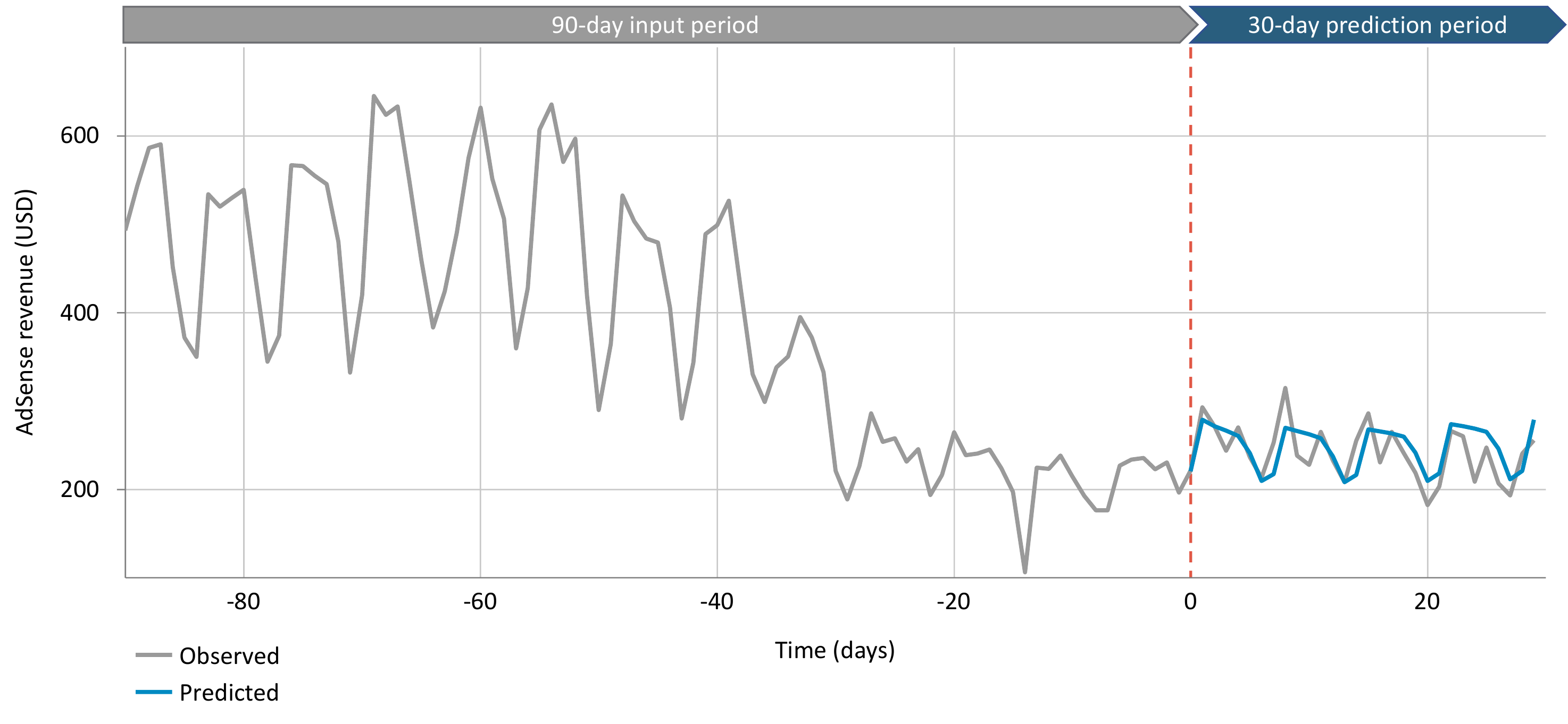}}
    \subfigure[TFT  - Worst 1 prediction (30 days)]{\includegraphics[width=0.48\textwidth, height=3.5cm]{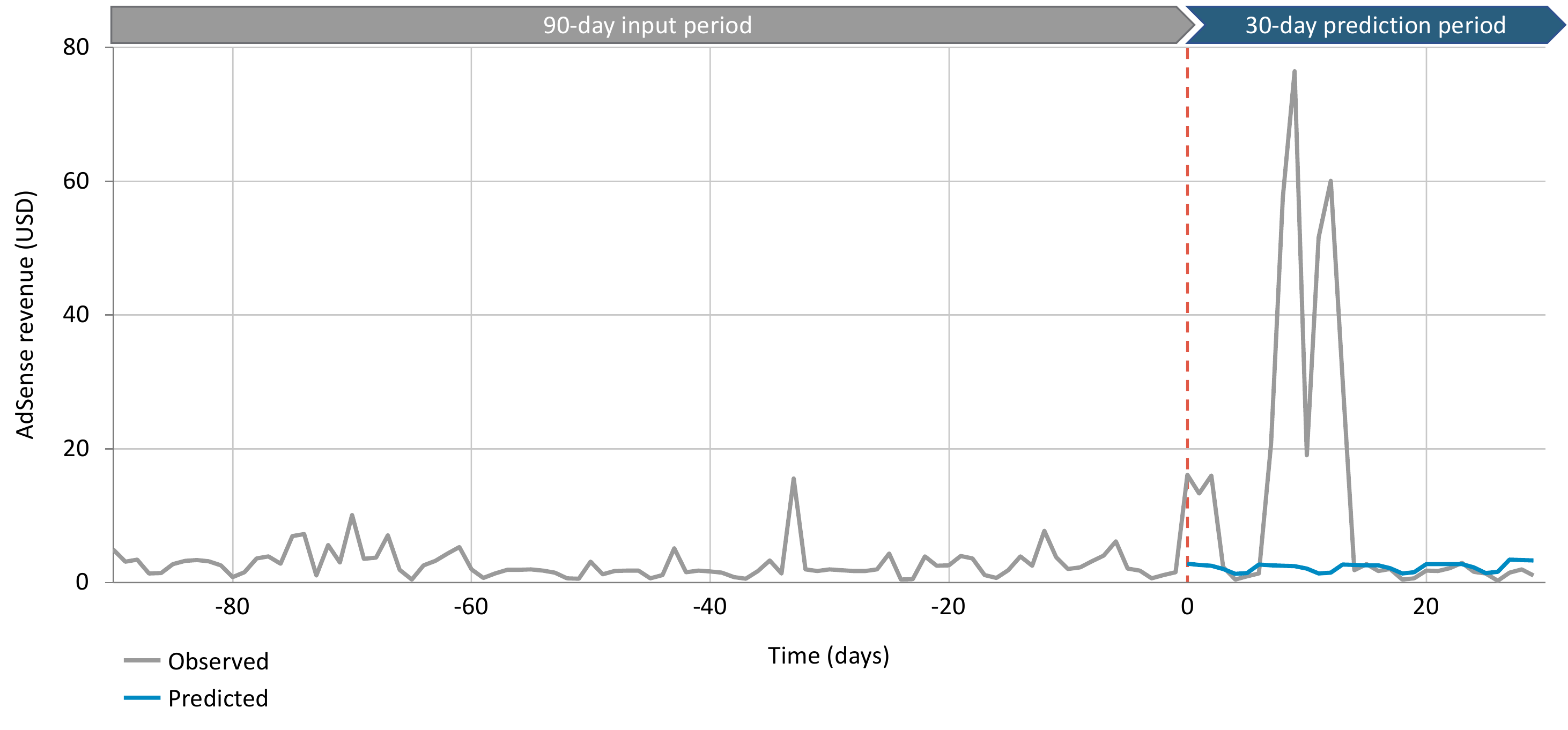}}
    \caption{The best (left) and worst (right) prediction of TFT model. The blue line indicates the actual revenue and orange line and band indicates the predicted revenue and its prediction intervals. Grey line indicates the attention weights measured on the right y-axis}
 \label{fig:top}
\end{figure*}

Further, we leverage the results from the TFT to generate insights into the underlying dynamics of the AdSense revenue predictions.
First, it is worth taking a look at the predictions generated by the model.
The best and worst TFT predictions are presented in Fig. \ref{fig:top}a and  Fig. \ref{fig:top}b.
On inspection of the best observation, it becomes clear that the TFT model picks up the weekly recurring patterns very well.
Furthermore, we note that our findings from Table I are confirmed as predictions are getting worse as the time horizon extends.
Looking at the worst prediction indicates the reason. We find that even sufficiently complex models like the TFT seem to occasionally struggle to compute sensible predictions.
While revenues remain flat during the input period, immediately after the prediction period starts, the revenue peaks, and the model provides very inaccurate results until the actual revenue series rebounds to historical levels.

\subsection{Feature importance}
\begin{figure*}[!htb]
    \centering
    \subfigure[Persistent weekly temporal Attention weight patterns ]{\includegraphics[width=0.43\textwidth, height=4cm]{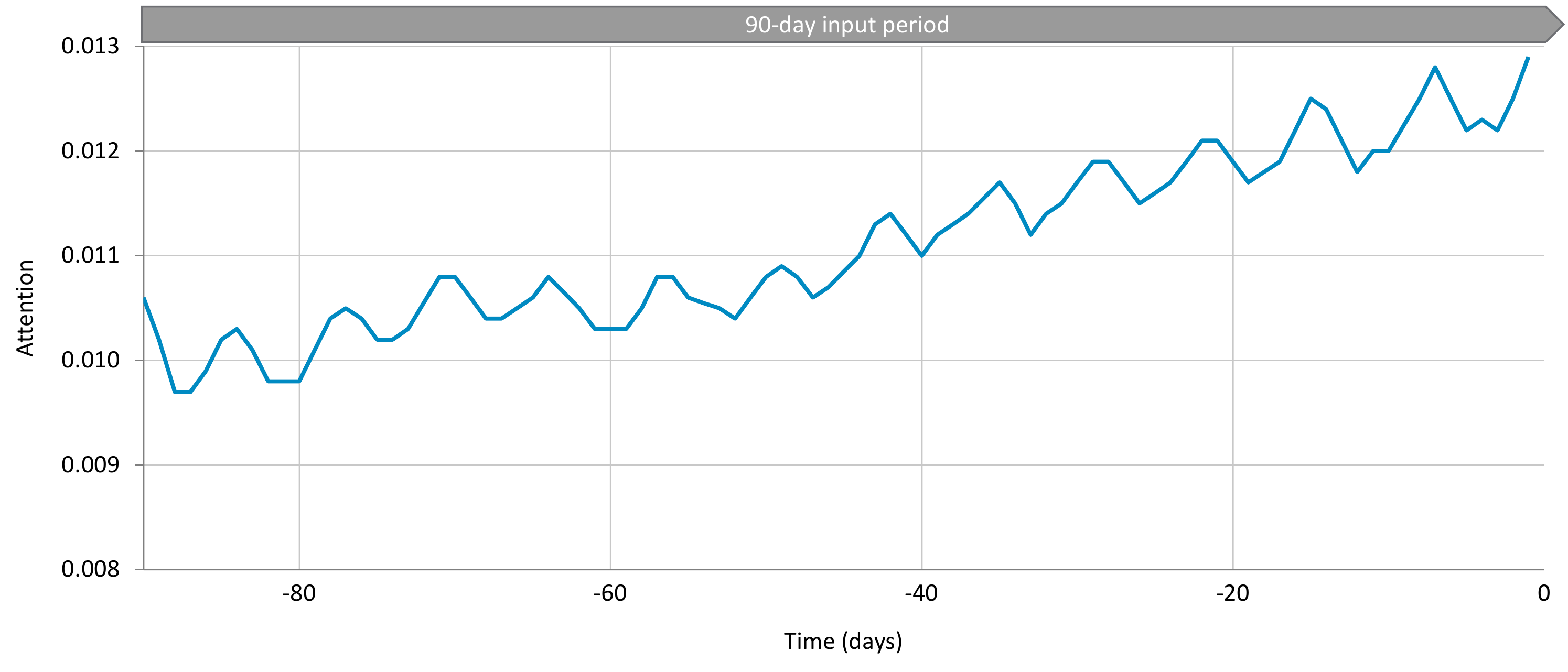}}
    \subfigure[Encoder Variable Importance ]{\includegraphics[width=0.55\textwidth, height=4cm]{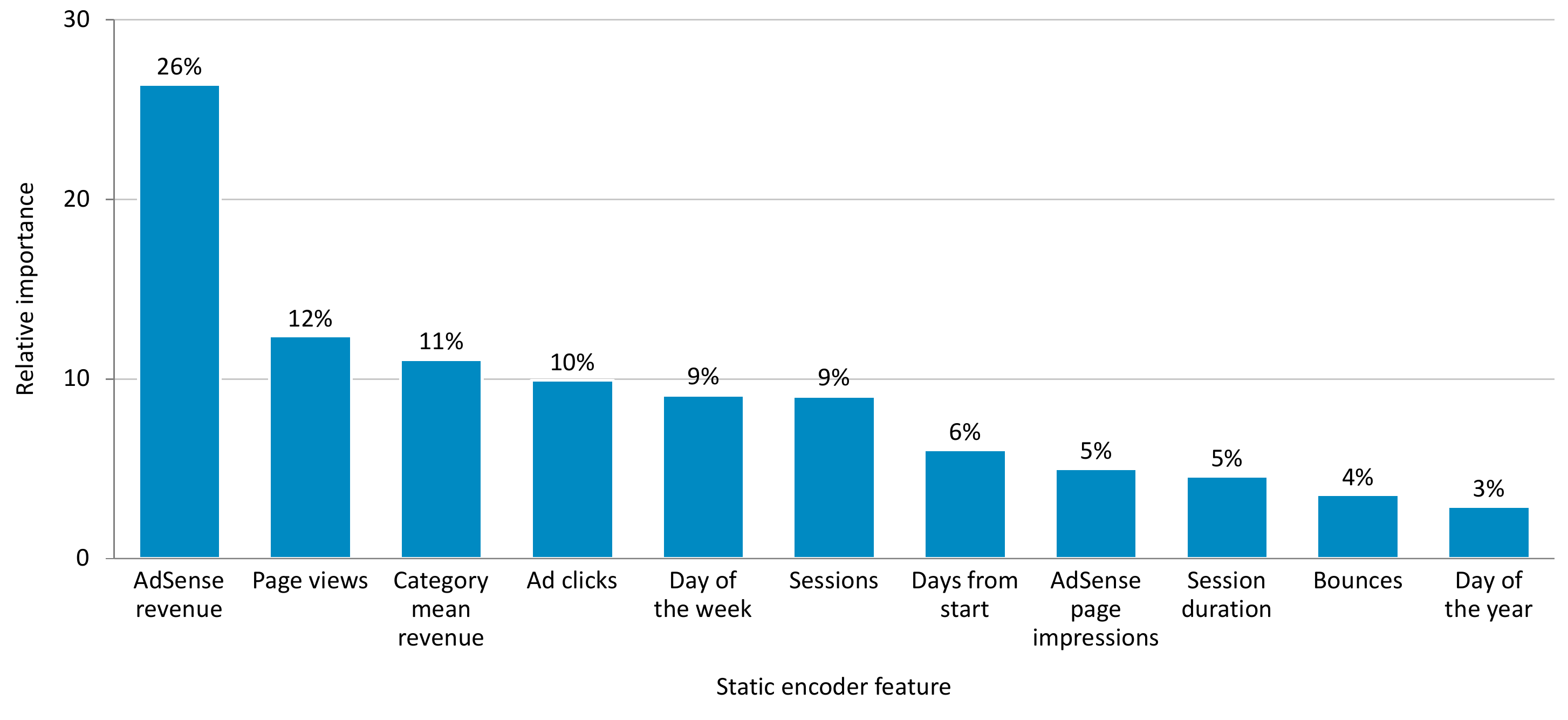}}
    \subfigure[Decoder Variable Importance ]{\includegraphics[width=0.48\textwidth, height=4cm]{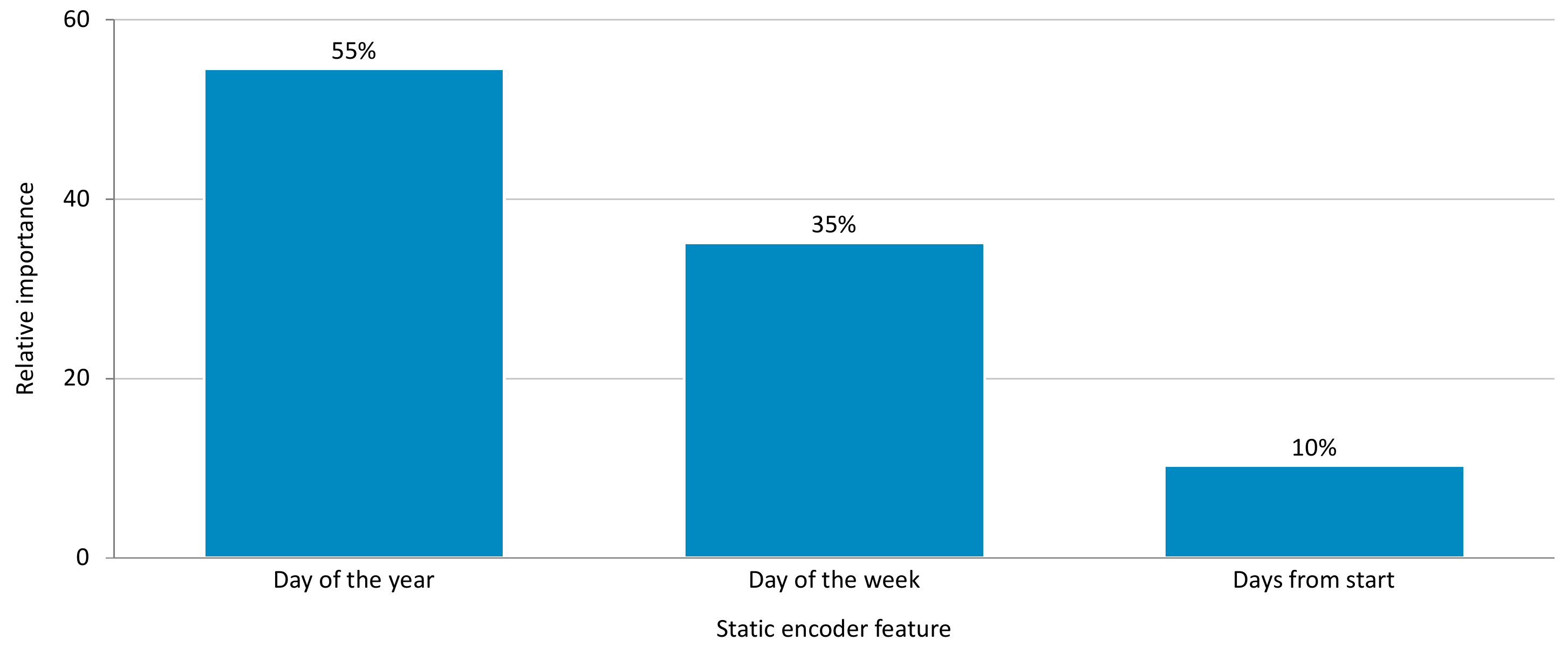}}
    \subfigure[Static Variable Importance ]{\includegraphics[width=0.48\textwidth, height=4cm]{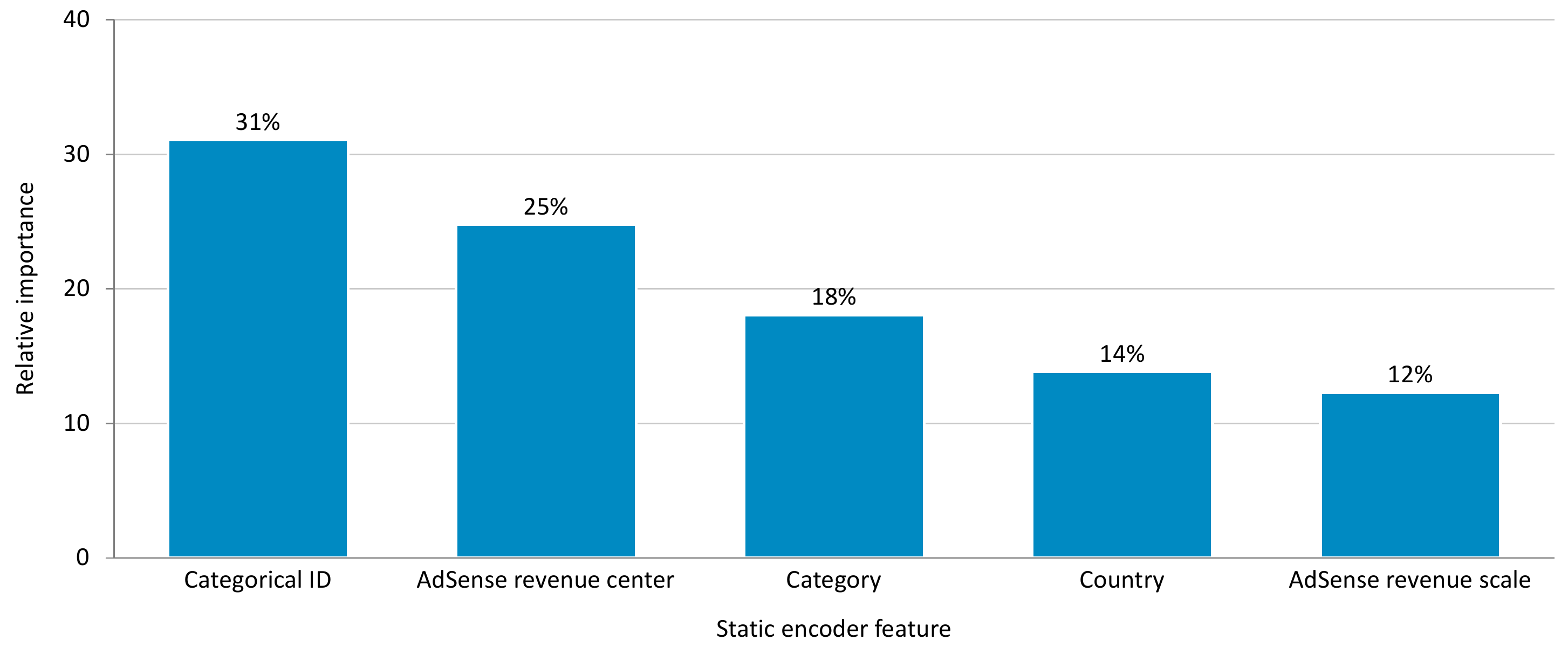}}
    \caption{TFT interpretability with Attention mechanism and variable importance in 30-day time horizon}
 \label{fig:tft}
\end{figure*}

Firstly, as elaborated in Sec. \ref{tft},  using the temporal attention weights from the trained TFT model, all attention heads share the same values such that overall attention can be computed by additive aggregation of the individual heads \cite{lim2020temporal}. More specifically, Fig. \ref{fig:tft}a exhibits the weekly recurring patterns of advertising revenues inherent in the data, which can be treated as evidence of the weekly web traffic patterns dominating advertising revenue development in the short term. This finding is consistent with periodic patterns of various online advertising statistics, including impressions, clicks, bids revealed in \cite{yuan2013}.

Feature importance is analyzed by interpreting the VSNs used by the TFT (Sec. \ref{tft}), which produce sparse weights for static, encoder, and decoder inputs.
Unsurprisingly, the encoder puts most of its weight on the advertising revenue (Fig. \ref{fig:tft}b).
Additionally, the day of the week and the category mean revenue dominate the encoder variable selection.
The importance of the week of the day, which was already observed in Fig. \ref{fig:top}a, can be interpreted as the first piece of evidence for what has been proposed on the basis of existing literature in Sec. \ref{online_ad_studies}, and the analysis of the autocorrelation plots in Sec. \ref{data}, that within a short-term prediction horizon, the weekly patterns associated with web traffic \cite{li2008forecasting} might yield the most significant determinant of advertising revenues.
Nevertheless, these findings do not oppose the results from previous literature (Sec. \ref{online_ad_studies}), indicating that there are three main determinants of advertising revenues, namely web traffic, CTR, and CPC.
Indeed, it should be noted that given the past ad impressions, ad clicks, and advertising revenues, the model can easily compute CTR and CPC as linear combinations.
But given the assumption that both CTR and CPC stay rather constant over the forecast period, they do not add much extra information to the revenue development as the model also has direct access to past advertising revenues.

Interestingly, the most important encoder variable next to the day of the week is the category mean revenue.
Thus, related advertising revenue time-series, like those of competitors, incorporate valuable information for making predictions.
This is in line with the previously mentioned performance improvement associated with the addition of the category mean revenue as an additional covariate.
Similar to the encoder, the decoder puts the most weight on the day of the year, followed by the day of the week (Fig. \ref{fig:tft}c), supporting the findings from the encoder VSNs. 

Finally, the static variable importance is dominated by the categorical id (Fig. \ref{fig:tft}d), which allows the model to uniquely identify any single publisher.
This observation indicates that the model cannot fit a single function that can generate predictions for all cross-sectional units. Hence, TFT still learns patterns that are associated only with a single publisher and which fail to generalize across publishers in the same group.
Such groupings could, for instance, be built on the basis of the country or category, which have also been treated as static features, but which turn out to be less important than the categorical id when making predictions.
Nevertheless, it is worth noting that the static VSN feature importance does not contradict the findings from \cite{DBLP:journals/corr/FlunkertSG17}, that a model may benefit from the access to cross-sectional data by being less prone to overfitting.

\section{Discussion and Conclusions}\label{discussion_conclusion}
In this paper, we perform an interpretable deep learning approach, namely the TFT model, to forecast publishers' online advertising revenues that outperforms not only many existing deep learning models but also reveals important features to interpret how and why the model makes such predictions. More specifically, we find that publishers' past advertising revenues, the mean advertising revenue of publishers in the same content category, and day of the week are the most important features when making predictions. Also, we find that the TFT mainly learns weekly temporal attention patterns, which are usually embedded in web traffic data. 

We also aim to take a step further in exploring related publishers' advertising revenues. Given multiple related online advertisement revenue time series, like those of competitors, the correlated time series might act as additional features in the prediction problem. Next to its business relevance, the prediction of advertising revenues for publishers constitutes a novel problem, which to our knowledge has not been investigated before. In particular, we show that model performance can be further improved when the past advertising revenue time-series from related publishers can be incorporated into the predictive model.
From a business perspective, this yields a unique competitive advantage for publishers who can have a holistic view of the competitive landscape of the online advertising market.

Our results provide strong managerial implications for publishers to consider improving advertising revenues. While CTR and  CPC  seem to be less volatile in the short term,  website traffic patterns directly transfer into short-term movements of advertising revenues. A sole focus through increasing revenue by optimizing CTR and CPR seems to be ineffective for practitioners. The main driver of revenue remains traffic to the page \cite{Aguilar2019}, but from the model importance, we can see that not all traffic is created equal. From the traffic-related metrics, the number of bounces has the strongest effect on the revenue, so optimizing the bounce rate is more important than just increasing the overall number of sessions. In return, the practice of some web pages to inflate the number of pageviews by photo galleries or similar techniques seems to have little effect, as the number of page views has the lowest impact on the model, especially in comparison to sessions and bounces. In conclusion, practitioners should focus on increasing the number of ``active" non-bounce sessions to maximize revenue.

Concerning deep learning applications in a business setting, this paper adds to the existing literature of time series forecasting in an underexplored area. The TFT model, given its VSNs and interpretable multi-head attention mechanism, offers a promising example of multi-step, multivariate time series forecasting, not only improving predictive performance but also interpretability. However, the results from this paper are only based upon a relatively small sample of publishers, where certain publisher even constitutes the only sample from a specific category. As the number of publishers increases, we may consider identifying more relevant correlated revenue time series, e.g., through temporal clustering techniques \cite{bandara2020}, and so we expect the significance of the results to improve further.





\bibliographystyle{IEEEtran}
\bibliography{bib.bib}

\end{document}